\documentclass[runningheads]{llncs}
\usepackage[T1]{fontenc}
\usepackage{wrapfig}
\usepackage{graphicx}
\usepackage{bbm}
\usepackage{makecell}
\usepackage{caption}
\usepackage{array}
\usepackage{multirow}
\usepackage{subcaption}
\usepackage{amssymb}
\usepackage{amsmath}
\usepackage{colortbl}
\usepackage{xcolor}
\usepackage{booktabs}

\definecolor{lightgray}{gray}{0.9}
\definecolor{linecolor}{rgb}{0.82, 0.94, 0.75}
\definecolor{evaunit01green}{RGB}{82,208,83}
\definecolor{lowred}{RGB}{238,18,137}

\definecolor{lowerred}{RGB}{255,110,180}

\begin{document}
\title{An Adapter-free Fine-tuning Approach for Tuning 3D Foundation Models}

\author{Sneha Paul\orcidID{0000-0001-7731-4196} \and
Zachary Patterson\orcidID{0000-0001-8878-7845} \and
Nizar Bouguila\orcidID{0000-0001-7224-7940}}

\authorrunning{S. Paul et al.}
%
\institute{Concordia University, Montreal, Canada \\}

\maketitle 
\begin{abstract}
Point cloud foundation models demonstrate strong generalization, yet adapting them to downstream tasks remains challenging in low-data regimes. Full fine-tuning often leads to overfitting and significant drift from pre-trained representations, while existing parameter-efficient fine-tuning (PEFT) methods mitigate this issue by introducing additional trainable components at the cost of increased inference-time latency.  
We propose \textbf{Momentum-Consistency Fine-Tuning (MCFT)}, an adapter-free approach that bridges the gap between full and parameter-efficient fine-tuning. MCFT selectively fine-tunes a portion of the pre-trained encoder while enforcing a momentum-based consistency constraint to preserve task-agnostic representations. Unlike PEFT methods, MCFT introduces no additional representation learning parameters beyond a standard task head, maintaining the original model’s parameter count and inference efficiency.
We further extend MCFT with two variants: a semi-supervised framework that leverages abundant unlabeled data to enhance few-shot performance, and a pruning-based variant that improves computational efficiency through structured layer removal. Extensive experiments on object recognition and part segmentation benchmarks demonstrate that MCFT consistently outperforms prior methods, achieving a 3.30\% gain in 5-shot settings and up to a 6.13\% improvement with semi-supervised learning, while remaining well-suited for resource-constrained deployment. 

\end{abstract}

\section{Introduction}
\label{sec:intro}

The recent rise of foundation models has showcased remarkable generalization capabilities and human-like performance across a range of domains. These models, pre-trained on large-scale datasets to learn general-purpose representations, have become essential tools in areas such as 2D computer vision and natural language processing (NLP). A similar trend is observed in point clouds, an unstructured 3D geometric data that plays a crucial role in real-world applications by enabling accurate object recognition and scene understanding in complex 3D environments. Foundation models for point clouds have shown promising performance and generalization capabilities \cite{pointmae,pointbert}. However, to fully leverage their potential for specific downstream tasks, additional fine-tuning is still required \cite{dgcnn,act}.

\begin{figure}
\centering
\begin{minipage}[b]{0.4\textwidth}
  \centering
  \includegraphics[width=0.8\linewidth]{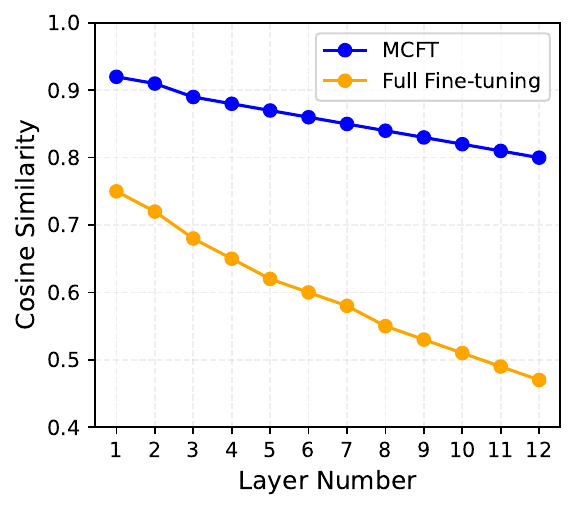}
  \caption{Layer-wise similarity}
  \label{fig:similarity}
\end{minipage}
\hspace{15pt}
\begin{minipage}[b]{0.5\textwidth}
  \centering
  \includegraphics[width=0.75\linewidth]{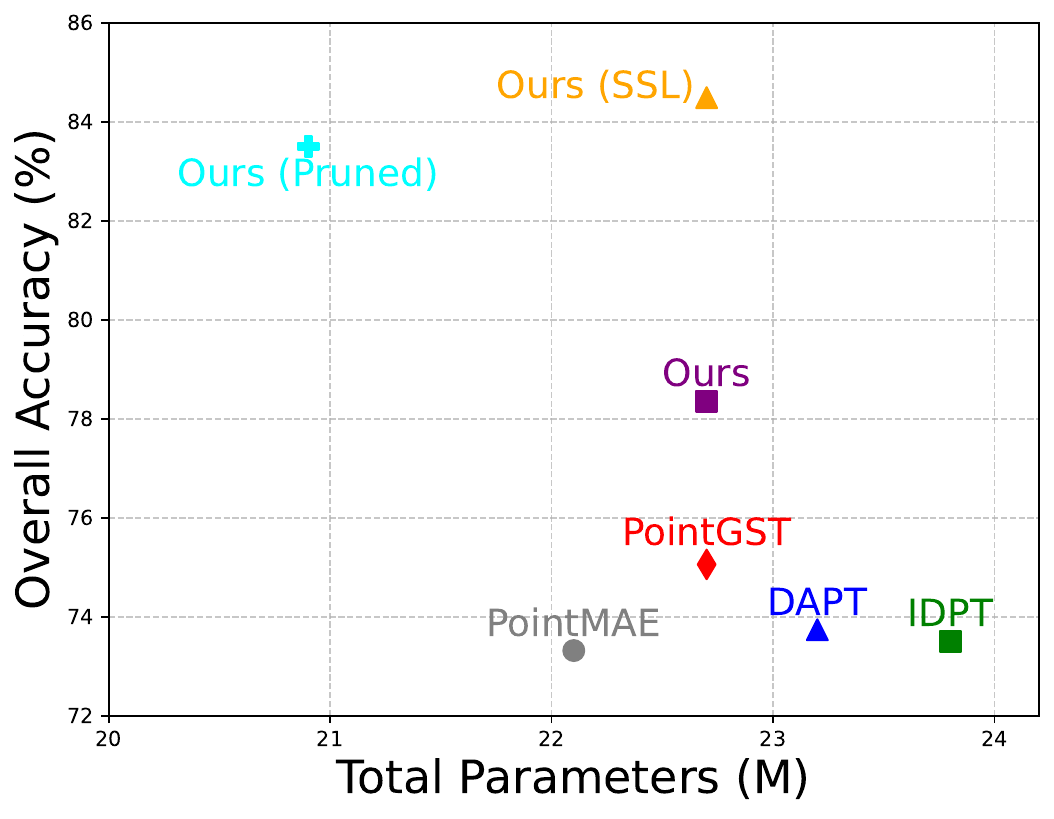}
  \caption{Parameters vs performance}
  \label{fig:banner}
\end{minipage}
\vspace{-15pt}
\label{fig:intro}
\end{figure}

Fine-tuning a foundation model for downstream tasks presents significant challenges, particularly when training data is limited. Tuning with scarce data often leads to overfitting and a loss of generalization, diminishing the foundation model’s effectiveness. For instance, as illustrated in Fig 1, full fine-tuning (FFT) leads to overfitting and significant divergence from the pre-trained model’s representations, particularly in deeper layers. In 2D vision and NLP domains, recent studies \cite{hulora,li2021prefix,chen2022adaptformer} have explored various Parameter-Efficient Fine-Tuning (PEFT) techniques to address this issue. These methods have achieved state-of-the-art (SOTA) performance by introducing additional trainable parameters, such as adapters \cite{chen2022adaptformer} or prompts \cite{coprompt}, while keeping the pre-trained model weights frozen. Inspired by this, PEFT techniques have been adopted to 3D as well \cite{idpt,dapt,pointgst}. However, unlike other domains, PEFT methods for 3D foundation models show only a minimal performance gain when applied in a low-data scenario (order of 1\% improvement), as illustrated in Fig 2. Furthermore, PEFT introduces additional trainable parameters to the frozen encoder, which increase the overall memory and compute cost of the model \textit{during inference} --- a bottleneck for deploying such a model in a resource-constrained environment like mobile devices. For instance, IDPT \cite{idpt} introduces 1.7 million extra parameters to the pre-trained PointMAE encoder (Fig 1), which results in a staggering 50\% increase in inference latency (see Table \ref{tab:classification} for more details).

To address these challenges, we propose Momentum-Consistency Fine-Tuning (MCFT), an adapter-free approach that bridges the gap between FFT and PEFT for fine-tuning 3D foundation models in low-data settings. Unlike PEFT methods that insert adapter modules or prompts into the frozen encoder for adaptation, MCFT does not introduce any additional parameters for tuning the encoder- except a standard task-specific prediction head (e.g. classifier layer).
The core idea of MCFT is a selective fine-tuning approach of a defined portion of the pre-trained encoder by enforcing our proposed consistency constant to avoid overfitting. This approach matches the original model's parameter count and computational efficiency without the additional representation learning parameters introduced by PEFT methods. As illustrated in Fig 2, MCFT maintains a high degree of similarity to the pre-trained encoder across layers, in contrast to FFT, which increasingly diverges, particularly in deeper layers, resulting in diminished generalization.

Additionally, we introduce two variants of MCFT to further improve performance and efficiency in few-shot scenarios. The first is a \textit{semi-supervised} momentum-consistency fine-tuning framework that leverages unlabelled data (which are easy and abundant to collect) alongside limited labelled data. 
Specifically, we adopt the semi-supervised learning protocol from AllMatch \cite{paul2024improving} in conjunction with our MCFT regularization to effectively utilize the unlabelled data.
The second variant is a lighter version of MCFT that focuses on enhancing computational efficiency through structural pruning. While MCFT already maintains the efficiency of the pre-trained encoder, this variant selectively prunes less important layers, producing a smaller and faster model without significantly compromising performance.

We evaluate MCFT extensively on object recognition in $n$-way few-shot \cite{zhang2025boosting}, full few-shot, and fully-supervised setups using synthetic \cite{modelnet} and real-world \cite{scanobjectnn} datasets, as well as part-segmentation tasks on ShapeNetPart \cite{shapenetpart}. Along with $n$-way (5 and 10-way) few-shot setups, we also report a full few-shot evaluation by testing performance across all dataset classes. This protocol is significantly more challenging than n-way evaluation as the model must learn to distinguish among all classes simultaneously rather than a small subset.
Our results show that MCFT consistently outperforms existing approaches in all settings. Notably, in the 5-shot evaluation, MCFT achieves a \textbf{3.30\%} improvement over prior methods, demonstrating its robustness in low-data scenarios. Furthermore, our semi-supervised variant yields an additional \textbf{6.13\%} increase over the standard few-shot results. Finally, our pruned variant maintains comparable performance to the full model while reducing computational overhead, making it ideal for resource-constrained environments.

\section{Related Work}
\label{related_work_suppl}

\textbf{PEFT in Point Clouds. }
Parameter-Efficient Fine-Tuning (PEFT) reduces adaptation cost by updating a small subset of parameters while keeping the backbone frozen. Common PEFT paradigms include prompt-based methods, which introduce learnable tokens into the input or attention space \cite{lester2021power,li2021prefix}, and adapter-based methods, which insert lightweight trainable modules into the network \cite{chen2022adaptformer}. Low-rank re-parameterization further improves efficiency by decomposing weight updates into compact subspaces \cite{hulora}. Recent extensions explore structural re-parameterization and task-specific token aggregation to better integrate task knowledge into frozen backbones \cite{luo2023towards,ruan2024gist}.
PEFT for 3D point clouds is relatively underexplored. IDPT \cite{idpt} first introduced instance-aware prompts for point cloud backbones, while DAPT \cite{dapt} combined dynamic adapters with prompt tuning. PPT \cite{ppt} proposed trainable positional embeddings derived from sampled and clustered tokens. PointPEFT \cite{pointpeft} integrates lightweight adapters with input-level prompts, achieving competitive performance with limited parameter overhead. However, existing methods uniformly introduce additional trainable parameters, increasing memory and training cost, which limits their suitability for deployment on resource-constrained devices.

\textbf{Semi-supervised Learning in Point Clouds.}
Semi-supervised learning (SSL) for point clouds aims to reduce reliance on labeled data. Prior work primarily focuses on enforcing prediction consistency across augmented point clouds \cite{chen2021consistency} or aligning features with class-level prototypes in multi-modal settings \cite{chen2021multimodal}. To mitigate class imbalance, ConFid \cite{chen2023class} resamples unlabeled data based on class-wise confidence, while AllMatch \cite{paul2024improving} maximizes utilization of the unlabeled set through unified matching strategies. Despite these advances, SSL has not been studied in the context of adapting large 3D foundation models.

\textbf{Pruning for Model Compression. }
Pruning improves model efficiency by removing parameters with low importance. Unstructured pruning achieves high sparsity but often requires specialized hardware \cite{sanh2020movement}, whereas structured pruning removes entire blocks or layers, yielding predictable acceleration \cite{lagunas2021block}. Pruning is commonly applied after fine-tuning, making it compatible with PEFT pipelines.
Recent methods combine pruning with low-rank adaptation or task-aware compression. SPA \cite{hedegaard2024structured} integrates structured pruning with compact adapters, while CPET \cite{zhao2023cpet} combines pruning, low-rank tuning, and distillation. Other approaches jointly prune and adapt parameters but rely on static tuning configurations \cite{zhang2024loraprune,li2022parameter}. APT \cite{zhao2024apt} addresses this limitation via dynamic salience-based tuning and pruning. However, pruning has not yet been explored for efficient adaptation of 3D foundation models.

\section{MCFT: Momentum-Consistency Fine-Tuning}
\label{sec:method}

Transformers have become an effective architecture for processing 3D point clouds due to their ability to capture complex spatial relationships~\cite{pointmae}. In a typical point cloud model, such as Point-MAE~\cite{pointmae}, a raw point cloud ${X} \in \mathbb{R}^{M \times 3}$, containing $M$ points, is segmented into smaller patches ${X}' \in \mathbb{R}^{m \times k \times 3}$ using Farthest Point Sampling (FPS) and K-Nearest Neighbour (KNN) algorithms. Here, $m$ denotes the number of patches, and each patch contains $k$ local points. These patches are transformed into input tokens, represented as ${E_0} \in \mathbb{R}^{m \times d}$,
where \(d\) is the embedding dimension. To incorporate positional information, the patches are further embedded using a point patch embedding module. Additionally, a classification token (\texttt{[CLS]}) is added to the sequence of input tokens. The combined sequence passes through a series of transformer layers, each consisting of self-attention and feed-forward modules, which are responsible for learning meaningful representations. 
Such an encoder, when pre-trained on large unlabelled data, learns generalizable features of a particular domain. Once pre-trained, the encoder can be fine-tuned with minimal supervision to perform any downstream task. However, fine-tuning a foundation model with limited labelled data is challenging, primarily due to the massive size of the model, which can easily lead to overfitting. To deal with such issues, PEFT incorporates additional learnable parameters into the frozen pre-trained model. While such a concept shows desirable performance improvement, it increases the number of parameters and the inference cost of the model. In this work, we focus on fine-tuning a foundation model to improve its performance on the downstream task without increasing its inference cost.

We propose Momentum-Consistency Fine-tuning (MCFT), an adapter-free tuning approach for tuning a foundation model without introducing any additional parameters \textbf{into the pre-trained encoder} to maintain the efficiency of the pre-trained encoder. Only a task-specific prediction layer (e.g. a classifier head) is added on top of the encoder to generate the prediction (e.g. class distribution), which is not specific to our method and is required to generate the prediction. The core idea of our proposed solution is to strategically fine-tune a few layers of the pre-trained encoder with a novel momentum-consistency-based self-distillation module that regularizes the learning process by aligning the embeddings of the trainable model (student) with those of the pre-trained model (teacher). We enable learning the new concept by fine-tuning a few layers of the pre-trained encoder; at the same time, the self-distillation concept ensures that the fine-tuning does not overfit the new small data and lose its generalization. However, a naive self-distillation (without our momentum update concept) with the pre-trained model restricts the learning to the knowledge of the pre-trained encoder. To alleviate this limitation, we slowly update the pre-trained encoder as the exponential moving average (EMA) of the learnable encoder. This provides a stable regularization target that prevents the learnable encoder from overfitting to limited training data.

Formally, let ${E}_{\text{s}} \in \mathbb{R}^{d}$ denote the [CLS] embedding from the student model and ${E}_{t} \in \mathbb{R}^{d}$ be the [CLS] embedding from the teacher model. The alignment loss is defined with a cosine distance loss function as:
\begin{equation}
\scriptsize
\mathcal{L}_{\text{align}} = 1 - \frac{{E}_{s}^T \cdot {E}_{t}}{\|{E}_{s}\| \|{E}_{t}\|},
\end{equation}
where $\mathcal{L}_{\text{align}}$ measures the cosine similarity between the [CLS] embeddings, promoting consistency across models. This alignment mitigates overfitting and ensures better generalization on downstream tasks. 

In addition, we \textit{add a classifier head} that takes the embedding ${E}_{s}$ and generates class predictions, $y_i$. This classifier head enables the model to directly learn task-specific features and improves classification performance. To train the classifier, we use a cross-entropy loss defined as follows: 
\begin{equation}
\scriptsize
    \mathcal{L}_{\text{sup}} = -\sum_{i=1}^C y_i \log(\hat{y}_i),
\end{equation}
where $C$ is the number of classes, $y_i$ is the ground-truth label, and $\hat{y}_i$ is the predicted probability for class $i$. Initially, the classifier head is trained along with the feature encoder for a fixed number of epochs. Then, we follow the full fine-tuning protocol where the parameters of the momentum encoder are updated using the EMA approach. Let, $\theta_s$ and $\theta_{t}$ represent the parameters of the online and momentum encoders, the EMA update is defined as $\theta_t = \alpha \cdot \theta_{t} + (1 - \alpha) \cdot \theta_{s}$,
where $\alpha \in [0,1]$ is the smoothing factor controlling the contribution of the previous parameters. The EMA mechanism ensures stable updates by gradually incorporating changes, preventing abrupt shifts in the model’s behaviour.
The final loss function combines the alignment and classification losses as:
$\mathcal{L} = \mathcal{L}_{\text{align}} + \lambda \cdot \mathcal{L}_{\text{sup}}$, where $\lambda$ is a weighting factor that balances the alignment and classification objectives.

\begin{table*}[!t]
  \centering
    \setlength{\tabcolsep}{1.5mm}
  \caption{Comparison of the object recognition task on the ModelNet40 and ScanObjectNN (OBJ\_ONLY) dataset in \textit{full} few-shot setup.}
  \resizebox{0.8\linewidth}{!}{
    \begin{tabular}{lcccccc}
    \toprule
       \multirow{2.3}{*}{Methods} & \multicolumn{3}{c}{ModelNet40} & \multicolumn{3}{c}{ScanObjectNN} \\
\cmidrule{2-7}   & 5-shot & 10-shot & 20-shot & 5-shot & 10-shot & 20-shot \\
    \midrule
   Point-BERT~\cite{pointbert} & 72.88$\pm$2.3  & 79.31$\pm$ 2.1 & 82.11$\pm$2.7 & 56.92$\pm$1.8  &  64.84$\pm$1.9 & 74.13$\pm$2.4 \\
   + IDPT~\cite{idpt}  & 73.97 $\pm$1.8& 80.58$\pm$ 1.9 & 84.43$\pm$1.7 & 57.23$\pm$2.1  &  65.88$\pm$2.5 & 75.11$\pm$2.3 \\
   + DAPT~\cite{dapt}  & 73.68$\pm$\textbf{1.3}  &  80.47$\pm$1.9 & 83.87$\pm$2.2 & 57.89$\pm$2.5  &  66.03$\pm$1.7 & 75.12$\pm$2.9 \\
   + PointGST~\cite{pointgst}  & 75.14$\pm$1.6  &  81.25$\pm$2.3 & 85.17$\pm$1.9 &59.35$\pm$1.9  &  66.14$\pm$1.8 & 76.08$\pm$\textbf{2.1 }\\
   \rowcolor{linecolor!40}+ \textbf{MCFT} &  \textbf{77.91}$\pm$2.3 &\textbf{82.93}$\pm$\textbf{1.2} &\textbf{86.68}$\pm$\textbf{1.7} &\textbf{60.97}$\pm$\textbf{1.2}  &  \textbf{67.23}$\pm$\textbf{1.5} & \textbf{76.78}$\pm$\textbf{2.1} \\
       \midrule
    Point-MAE~~\cite{pointmae} & 73.32$\pm$2.1  & 79.97$\pm$ 2.5 & 83.78$\pm$1.7 & 56.37$\pm$1.2  &  65.39$\pm$1.8 & 74.86$\pm$2.2 \\
   + IDPT~\cite{idpt}  & 73.51 $\pm$2.8& 80.24$\pm$ 1.7 & 83.91$\pm$1.6& 57.01$\pm$2.2  &  65.88$\pm$2.5 & 75.58$\pm$2.7 \\
   + DAPT~\cite{dapt}  & 73.74$\pm$1.6  &  80.83$\pm$2.1 & 84.40$\pm$1.9 &57.21$\pm$2.2  &  66.25$\pm$2.6 & 75.81$\pm$2.1 \\
   + PointGST~\cite{pointgst}  & 75.06$\pm$1.9  &  81.02$\pm$2.9 & 84.91$\pm$2.7 &58.23$\pm$1.8  &  66.75$\pm$2.2 & 76.18$\pm$\textbf{1.1 }\\
   \rowcolor{linecolor!40}+ \textbf{MCFT} &  \textbf{78.36}$\pm$\textbf{0.5}  &  \textbf{82.45}$\pm$\textbf{1.0} & \textbf{86.83}$\pm$\textbf{1.3} &
   \textbf{61.10}$\pm$\textbf{0.9}  &  \textbf{67.98}$\pm$\textbf{1.2} & \textbf{77.45}$\pm$\textbf{1.1} \\
    \bottomrule
    \vspace{-20pt}
    
    \end{tabular}
  }
  \label{tab:full_fewshot}
\end{table*}

{Building on the core MCFT framework, we further propose a semi-supervised variant of MCFT, that leverages unlabelled data along with the limited labelled data to improve the performance.}
Let \( X_{lb} = \{(x_i, y_i)_{i=1}^{n}\} \) and \( X_{ulb} = \{(x_i)_{i=1}^{N}\} \) represent the labelled and unlabelled datasets, with \( n \ll N \). The learning objective minimizes both supervised loss \( \mathcal{L}_{sup} \) and unsupervised loss \( \mathcal{L}_{unsup} \) as:
\begin{equation}
\scriptsize
    \min_{\theta} \Big[
    \sum_{(x_i, y_i) \in X_{lb}} \mathcal{L}_{sup}(x_i, y_i, \theta) 
    + \omega \sum_{x_i \in X_{ulb}} \mathcal{L}_{unsup}(x_i, \theta)
    \Big],
    \label{equ:semiLoss}
\end{equation}
\vspace{-10pt}
\begin{equation}
\scriptsize
\mathcal{L}_{em} = \frac{1}{\mu B} \sum_{b=1}^{\mu B} \mathbbm{1} \left( \max(p_m(p_z(x^{(i)}_w))) \geq \tau \right) \cdot H \left( \arg\max(p_m(p_z(x^{(i)}_w))), p_m(p_z(x^{(i)}_s)) \right),
\label{eq:loss_unsup}
\end{equation}
\vspace{5pt}

\begin{wraptable}{r}{0.5\textwidth}
\vspace{-20pt}
  \centering
  \scriptsize
    \setlength{\tabcolsep}{0.7mm}
  \caption{Performance on ModelNet40~\cite{modelnet} in \textit{n-way} few-shot setup.}
  \resizebox{0.99\linewidth}{!}{
    \begin{tabular}{lccccc}
    \toprule
   \multirow{2.3}{*}{Methods} & \multicolumn{2}{c}{5-way} & \multicolumn{2}{c}{10-way} \\
\cmidrule{2-5}  &        10-shot & 20-shot & 10-shot & 20-shot \\
    \midrule
    OcCo~\cite{occo} & 94.0$\pm$3.6& 95.9$\pm$2.3 & 89.4$\pm$5.1 & 92.4$\pm$4.6 \\
    Point-BERT~\cite{pointbert}  & 94.6$\pm$3.1 & 96.3$\pm$2.7 & 91.0$\pm$5.4 & 92.7$\pm$5.1 \\
    MaskPoint~\cite{maskpoint} & 95.0$\pm$3.7 & 97.2$\pm$1.7 & 91.4$\pm$4.0 & 93.4$\pm$3.5 \\
    Point-MAE~\cite{pointmae} & 96.3$\pm$2.5 & 97.8$\pm$1.8 & 92.6$\pm$4.1 & 95.0$\pm$3.0 \\
    Point-M2AE~\cite{point2mae}   & 96.8$\pm$1.8 & 98.3$\pm$1.4 & 92.3$\pm$4.5 & 95.0$\pm$3.0 \\
    ACT~\cite{act} & 96.8$\pm$2.3 & 98.0$\pm$1.4 & 93.3$\pm$4.0 & 95.6$\pm$2.8 \\
    RECON~\cite{recon}  & 97.3$\pm$1.9 & 98.9$\pm$3.9 & 93.3$\pm$3.9 & 95.8$\pm$3.0 \\
    \midrule
   Point-BERT~\cite{pointbert}   &94.6$\pm$3.1 & 96.3$\pm$2.7 & 91.0$\pm$5.4 & 92.7$\pm$5.1 \\
   + IDPT~\cite{idpt}    & 96.0$\pm$\textbf{1.7}& 97.2$\pm$2.6& 91.9$\pm$4.4& 93.6$\pm$3.5\\
  + DAPT~\cite{dapt} & 95.8$\pm$2.1 &97.3$\pm$1.3 &92.2$\pm$4.3 &94.2$\pm$3.4 \\
  + PointGST~\cite{pointgst}  & \textbf{96.5}$\pm$2.4  &  \textbf{97.9}$\pm$2.0 & 92.7$\pm$4.2 & \textbf{95.0}$\pm$2.8  \\
   \rowcolor{linecolor!40}+ \textbf{MCFT} & \textbf{96.5}$\pm$1.9 &97.6$\pm$\textbf{1.1} &\textbf{93.5}$\pm$\textbf{2.9} &\textbf{95.0}$\pm$\textbf{1.7} \\
       \midrule
    Point-MAE~~\cite{pointmae}   & 96.3$\pm$2.5 & 97.8$\pm$1.8 & 92.6$\pm$4.1 & 95.0$\pm$3.0\\
   + IDPT~\cite{idpt}  & 97.3$\pm$2.1& 97.9$\pm$1.1& 92.8$\pm$4.1& 95.4$\pm$\textbf{2.9}\\
   + DAPT~\cite{dapt}  & 96.8$\pm$1.8  &  98.0$\pm$1.0 & 93.0$\pm$3.5 & 95.5$\pm$3.2  \\
   + PointGST~\cite{pointgst}  & 98.0$\pm$1.8  &  \textbf{98.3}$\pm$\textbf{0.9} & 93.7$\pm$4.0 & 95.7$\pm$2.4  \\
   \rowcolor{linecolor!40}+ \textbf{MCFT} & \textbf{98.1}$\pm$\textbf{1.0}  &  \textbf{98.3}$\pm$1.2 & \textbf{94.3}$\pm$\textbf{2.2} & \textbf{95.9}$\pm$\textbf{2.9}  \\
    \bottomrule
    \end{tabular}
    }
      \label{tab:fewshot}
    \vspace{-20pt}
\end{wraptable}
For unsupervised learning, we adopt the concept of entropy minimization by using pseudo-labelling, where high-confidence predictions from weakly augmented samples serve as supervision for their strongly augmented counterparts. Inspired by FixMatch \cite{fixmatch}, the unsupervised loss is defined in Eq. \ref{eq:loss_unsup}.

where \( \tau \) is the confidence threshold, \( \mu \) represents the ratio of unlabelled to labelled batch sizes, and \( H \) denotes the cross-entropy loss. To address the limitations of static thresholds and ensure the inclusion of all unlabelled samples, we adopt the semi-supervised framework from AllMatch ($\mathcal{L}_{AllMatch}$) \cite{paul2024improving}. This framework comprises three modules: adaptive hard augmentation, inverse learning, and contrastive learning, which collectively enhance the use of unlabelled data to improve generalization, particularly when labelled data is limited. Finally, our unsupervised learning is the combination of the entropy minimization loss and alignment loss: 
\begin{equation}
\scriptsize
    \mathcal{L}_{\text{unsup}} = \mathcal{L}_{\text{align}}+\mathcal{L}_{AllMatch}
\end{equation}

\vspace{-5pt}
\begin{wraptable}{r}{0.5\textwidth}
\vspace{-15pt}
  \centering
  \scriptsize
    \setlength{\tabcolsep}{0.4mm}
  \caption{Performance comparison of object recognition task in \textit{fully-supervised} setting.
  }
  
  \vspace{-5pt}
  \resizebox{0.99\linewidth}{!}{
    \begin{tabular}{lcccccccc}
    \toprule
    \multirow{2.3}{*}{Method} &\multirow{2.3}{*}{\#TP (M)} &\multirow{2.3}{*}{FLOPs} &\multicolumn{3}{c}{ScanObjectNN} &\multicolumn{1}{c}{ModelNet40}\\
		\cmidrule(r){5-7} \cmidrule{8-9}
	& & & BG & ONLY &PB  & OA (\%)      \\
    \midrule
    PointNet~\cite{pointnet} & 3.5 & 0.5  & 73.3  & 79.2  & 68.0 &   89.2 \\
    PointNet++~\cite{pointnet++}   & 1.5 & 1.7 & 82.3  & 84.3  & 77.9 &  90.7\\
    DGCNN~\cite{dgcnn}  & 1.8 & 2.4 & 82.8  & 86.2  & 78.1 &   92.9 \\
    PointMLP~\cite{pointmlp}  &  13.2 & 31.4  & -    & -     & 85.4   & 94.5\\
    \midrule
    OcCo~\cite{occo} & 22.1 & 4.8 & 84.9 & 85.5 & 78.8  & 92.1 \\
    Point-BERT~\cite{pointbert} & 22.1 & 4.8  & 87.4 & 88.1 &  83.07  &  93.2 \\
    MaskPoint~\cite{maskpoint} & 22.1 & - & 89.7 & 89.3 &  84.6 &   93.8 \\
    Point-MAE~\cite{pointmae} & 22.1 & 4.8 & 90.0 & 88.3 & 85.2 &  93.8 \\
    Point-M2AE~\cite{point2mae} & 15.3 & 3.6 & 91.2 & 88.8 & 86.4 & 94.0\\
    ACT~\cite{act} & 22.1 & 4.8 & 93.3 & 91.9  & 88.2 &  93.7\\
    RECON~\cite{recon} & 43.6 & 5.3 & 94.2 & 93.1  & 89.7 &   93.9  \\
    \midrule
    Point-BERT~\cite{pointbert}  & 22.1 & 4.8 & 87.4 & 88.1 & 83.1&  93.2\\
    + PointPEFT \cite{pointpeft}& 22.8 & -- & {--} & {--} & {85.0} & --\\
    + IDPT~\cite{idpt} & 23.8 & 7.2 & {88.1}  & {88.3}  & {83.7}  &  93.4\\
    + DAPT \cite{dapt}& 23.2 & 5.0 & {91.1} & {89.7} & {85.4} & 93.6\\
    + PointGST \cite{pointgst}& \textbf{22.7} & \textbf{4.8} & {91.4} & {89.7} & {85.6} & 93.8\\
    \rowcolor{linecolor!40}+ \textbf{MCFT}  &  \textbf{22.7} & \textbf{4.8} & \textbf{{92.1}} & \textbf{{91.5}} & \textbf{{88.3}}  &  \textbf{95.2}\\
    \midrule
    Point-MAE~\cite{pointmae} & 22.1 & 4.8& 90.0 & 88.3 & {85.2}  & 93.8\\
    {\color[gray]{0.8} + PPT} \cite{ppt} & {\color[gray]{0.8} 23.4} & {\color[gray]{0.8} 4.8} & {\color[gray]{0.8} 92.9} & {\color[gray]{0.8} 92.4} & {\color[gray]{0.8} 88.4} & {\color[gray]{0.8} 93.4} \\
    {\color[gray]{0.8} + MoST} \cite{most} & {\color[gray]{0.8} 23.4} & {\color[gray]{0.8} 4.8} & {\color[gray]{0.8} --} & {\color[gray]{0.8} --} & {\color[gray]{0.8} 91.9} & {\color[gray]{0.8} --} \\
    + IDPT~\cite{idpt} & 23.8 & 7.2 & {91.2} & {90.0}& 84.9  &  94.4 \\
    + DAPT \cite{dapt} &  23.2 & 5.0 & {90.9} & {90.2} & {85.1}  &  94.0\\
    + PointGST \cite{pointgst} &  \textbf{22.7} & \textbf{4.8} & {91.7} & {90.2} & {85.3}  &  94.0\\
    \rowcolor{linecolor!40}+ \textbf{MCFT}  &  \textbf{22.7} & \textbf{4.8} & \textbf{{91.8}} & \textbf{{90.9}} & \textbf{{87.2}}  &  \textbf{94.9}\\
    \midrule
    RECON~\cite{recon}  & 22.1 & 4.8 & {94.3} & {92.8} & {90.0}&  93.0\\
    {\color[gray]{0.8} + PPT} \cite{ppt} & {\color[gray]{0.8} 23.2} & {\color[gray]{0.8} --} & {\color[gray]{0.8} 94.1} & {\color[gray]{0.8} 93.2} & {\color[gray]{0.8} 88.9} & {\color[gray]{0.8} 93.4} \\
    {\color[gray]{0.8} + MoST} \cite{most} & {\color[gray]{0.8} 23.4} & {\color[gray]{0.8} 4.8} & {\color[gray]{0.8} --} & {\color[gray]{0.8} --} & {\color[gray]{0.8} 92.8} & {\color[gray]{0.8} 94.7} \\
    + IDPT ~\cite{idpt}  & 23.8  & 7.2 & 93.3  &91.6  & 87.3  &  93.5 \\ 
    + DAPT~\cite{dapt}  & 23.2 & 5.0 &{94.3}  & {92.4}  & {89.4}  &   94.1\\
    + PointGST \cite{pointgst} &  \textbf{22.7} &  \textbf{4.8} &  94.5 & 92.9 &  89.5 &  94.1 \\
    \rowcolor{linecolor!40}+ \textbf{MCFT}  & \textbf{22.7} & \textbf{4.8} &\textbf{{94.9}} & \textbf{{93.1}}  & \textbf{90.8}  &   \textbf{95.2}\\
    \bottomrule
    \end{tabular}
    }
      \label{tab:classification}
      \vspace{-5pt}
\end{wraptable}
To further enhance MCFT's efficiency, we utilize a pruning technique that employs a salience scoring function to learn pruning masks \cite{zhao2024apt}.
This pruned version of MCFT improves the computational efficiency of transformer-based models for 3D point cloud processing, utilizing a pruning technique that employs a salience scoring function to learn pruning masks \cite{zhao2024apt}. These masks identify and remove irrelevant blocks of model parameters, thereby reducing computational overhead during inference.

Let \(\Theta\) denote the set of parameters of the transformer model, which consists of \(N\) layers. We define a pruning mask \(m \in \{0, 1\}^N\), where \(m_i = 1\) indicates that layer \(i\) is retained and \(m_i = 0\) indicates that it is pruned. The salience score for each layer \(i\) is computed based on the gradient of the loss \(L\) with respect to the output \(E_i\):
\begin{equation}
\vspace{-5pt}
\scriptsize
s_i = \frac{\partial L}{\partial E_i}
\vspace{-5pt}
 \end{equation}

\begin{wraptable}{r}{0.25\linewidth}
\vspace{-10pt}
\scriptsize
\setlength{\tabcolsep}{-1.mm}
\centering
\caption{Performance of different PEFT methods.
}
\vspace{-5pt}
\label{tab:peft_compare}
\resizebox{0.99\linewidth}{!}{
\begin{tabular}{ lcc }
\toprule
 Method & PB\_T50\_RS \\
\midrule
 Point-MAE~\cite{pointmae}   &  85.18  \\
 Linear probing &  75.99\\
 \midrule
  + Adapter~\cite{houlsby2019parameter}&  83.93 \\
  + Prefix tuning~\cite{li2021prefix} & 77.72  \\
  + BitFit~\cite{zaken2022bitfit}   & 82.62    \\
  + LoRA~\cite{hulora}  &   81.74   \\
  + VPT-Deep~\cite{jia2022visual}  &  81.09 \\
  + AdaptFormer~\cite{chen2022adaptformer}   & 83.45 \\
  + SSF~\cite{lian2022scaling}     & 82.58\\
  + IDPT~\cite{idpt}   &84.94\\
  + DAPT~\cite{dapt}   & 85.08\\
  + PointGST~\cite{pointgst}   & 85.29\\
  \rowcolor{linecolor!40}+ \textbf{MCFT}  & \textbf{87.21} \\
\bottomrule
\vspace{-15pt}
\end{tabular}
}
\end{wraptable}
To obtain the salience score, we first calculate the mean and standard deviation of the salience scores across layers:
 \begin{equation}
 \scriptsize
\mu_s = \frac{1}{N} \sum_{j=1}^N s_j, \quad \sigma_s = \sqrt{\frac{1}{N} \sum_{j=1}^N (s_j - \mu_s)^2}
\vspace{-5pt}
 \end{equation}
The final salience score for each layer is normalized as:
\begin{equation}
\scriptsize
\tilde{s}_i = \frac{s_i - \mu_s}{\sigma_s + \epsilon}
\vspace{-5pt}
\end{equation}
where \(\epsilon\) is a small constant to prevent division by zero. Next, we mask the desired number of layers based on our pruning budget. The pruning process is iteratively applied every \(K\) epoch during fine-tuning, allowing the model to adapt to the changes in layer importance until we reach our desired pruning budget. Specifically, after every \(K\) epoch, the salience scores are recalculated, and the pruning mask is updated accordingly. 

\vspace{-5pt}
\section{Experiments}
\vspace{-5pt}

\textbf{Few-shot learning.} In the full-few-shot setup, we evaluate MCFT on 5-, 10-, and 20-shot scenarios on both the ModelNet40 and ScanObjectNN datasets (Table \ref{tab:full_fewshot}), using PointMAE and PointBERT pre-trained encoders. For a fair comparison, we reproduce the performance of existing SOTA methods under identical experimental setups. Our experiments show that MCFT consistently outperforms the existing SOTAs across all few-shot scenarios. The largest improvement of \textit{3.30\%} is obtained in the 5-shot setting, where data is most limited.  
A similar result is also seen in 10- and 20-shot settings with the PointMAE encoder.
With the PointBERT encoder, MCFT continues to outperform SOTA by a significant margin in all few-shot settings. A similar trend is also observed on the ScanObjectNN dataset.

We conduct an additional few-shot analysis following the conventional $n$-way $m$-shot setup commonly used in previous works \cite{idpt,dapt,pointgst}. Table \ref{tab:fewshot} compares our method with existing approaches on 
5-way 10-shot, 5-way 20-shot, 10-way 10-shot, and 10-way 20-shot settings on the ModelNet40 dataset. These settings represent relatively simpler few-shot tasks, where specific subsets of classes are selected for training. Even in these near-saturation conditions, MCFT outperforms SOTA, reinforcing its effectiveness within the few-shot learning paradigm.

\begin{wraptable}{r}{0.3\linewidth}
\vspace{-20pt}
  \centering
  \scriptsize
  \setlength{\tabcolsep}{1.5mm}
  \caption{Performance on part segmentation task on ShapeNetPart.}
    \vspace{-5pt}
    \resizebox{0.99\linewidth}{!}{
    \begin{tabular}{lcc}
    \toprule
    &  \multicolumn{2}{c}{mIoU}\\
    Methods & Cls. &Inst. \\
    \midrule
    PointNet \cite{pointnet}  &80.4 & 83.7 \\
    PointNet++  \cite{pointnet++}     & 81.9 & 85.1 \\
    DGCNN \cite{dgcnn}   & 82.3 & 85.2 \\
    \midrule
    MaskPoint \cite{maskpoint}   & 84.6 & 86.0 \\
    Point-BERT \cite{pointbert}   & 84.1 & 85.6 \\
    Point-MAE \cite{pointmae}   & 84.2 & 86.1 \\ 
    ACT \cite{act}   & 84.7 & 86.1 \\
    \midrule
    Point-BERT~\cite{pointbert}    & 84.1 & 85.6 \\ 
    + IDPT~\cite{idpt}    & 83.5  & 85.3  \\
    + PointPEFT \cite{pointpeft}   &81.1 & 84.3\\
    + DAPT \cite{dapt}   & 83.8 & 85.5 \\
    + PointGST \cite{pointgst}   &83.9 & 85.7\\
    \rowcolor{linecolor!40}+ \textbf{MCFT}    & \textbf{85.2} & \textbf{87.1}\\
    \midrule
    Point-MAE~\cite{pointmae}    & 84.2 & 86.1 \\  
    + IDPT~\cite{idpt}    & 83.8  & 85.7  \\
    + PointPEFT \cite{pointpeft}   &83.2 & 85.2\\
    + DAPT \cite{dapt}  & 84.0 & 85.7 \\
    + PointGST \cite{pointgst}   &83.8 & 85.8\\
    \rowcolor{linecolor!40}+ \textbf{MCFT}    & \textbf{86.4} & \textbf{89.0}\\
    \midrule
    ReCon \cite{recon}   & 84.5 & 86.1 \\

    + IDPT~\cite{idpt}  & 83.7  & 85.7 \\
    + PointPEFT \cite{pointpeft} &83.1 & 85.1\\
    + DAPT \cite{dapt}   &83.9 & 85.7\\
    + PointGST \cite{pointgst}  &84.0 & 85.8\\
    \rowcolor{linecolor!40}+ \textbf{MCFT}    & \textbf{84.4 }& \textbf{86.8}\\
    \bottomrule
    \end{tabular}
    }
     \vspace{-20pt}
  \label{tab:segmentation}
\end{wraptable} 
\textbf{Fully-supervised learning.} In Table \ref{tab:classification}, we evaluate our proposed method on ModelNet40 and three variants of the ScanObjectNN dataset --- OBJ\_BG, OBJ\_ONLY, and PB\_T50\_RS, each with distinct levels of complexity, following the evaluation protocol of most existing methods. 
For a fair comparison, we follow the reported results of existing methods, with Point-BERT, Point-MAE, and RECON as pre-trained models, as provided by \cite{pointgst}. As we observe from Table \ref{tab:classification}, MCFT outperforms existing SOTA by 0.7\%, 1.8\%, and 2.7\% on the three variants of ScanObjectNN with Point-BERT encoder, while having similar FLOPs and total parameters to the pre-trained encoder. 
Similar trends are observed with Point-MAE and RECON as a backbone, underscoring the efficiency of MCFT. On the ModelNet40 dataset, though the performance of existing models is near saturation, our method still demonstrates consistent performance gain over SOTA methods, with a 1.4\% improvement for the Point-BERT encoder.

In Table \ref{tab:peft_compare}, we compare MCFT with several existing  PEFT techniques originally developed for NLP and vision tasks. These results, also adopted from \cite{dapt}, were performed on the PB\_T50\_RS variant of ScanObjectNN, the most challenging variant. PointMAE is used as the pre-trained encoder for this experiment. While most of the existing PEFT methods typically introduce additional parameters during training, increasing computational overhead during inference, MCFT outperforms all the existing PEFT methods with substantial margins without introducing additional parameters.

\begin{wraptable}{r}{0.3\linewidth}
\vspace{-25pt}
  \centering
  \scriptsize
  \setlength{\tabcolsep}{0.5mm}
  \caption{Performance on semi-supervised setting on ModelNet40.}
    \vspace{-5pt}
    \resizebox{0.99\linewidth}{!}{
    \begin{tabular}{lc}
    \toprule
       {Methods} &  5-shot \\
    \midrule
    Point-MAE~~\cite{pointmae} & 73.32$\pm$2.1 \\
   + IDPT~\cite{idpt}  & 73.51 $\pm$2.8\\
   + DAPT~\cite{dapt}  & 73.74$\pm$1.6 \\
   + PointGST~\cite{pointgst}  & 75.06$\pm$1.9 \\
   \rowcolor{linecolor!40}+ MCFT & 78.36$\pm$\textbf{0.5} \\
   \rowcolor{linecolor!40}+ \textbf{MCFT (SSL)} & \textbf{84.49}$\pm$0.7 \\
    \bottomrule
    \end{tabular}
    }
    \vspace{-20pt}
  \label{tab:ssl_fewshot}
\end{wraptable}
\textbf{Part segmentation.} To assess MCFT's performance in a dense prediction task, we evaluate it on the more challenging problem of part segmentation, where the goal is to predict a semantic label for each point in a 3D shape. Following prior work \cite{dapt}, we benchmark MCFT on the ShapeNetPart dataset \cite{shapenetpart}. As shown in Table \ref{tab:segmentation}, MCFT surpasses SOTA methods by a significant 2.6 and 3.2 mIoU in both class-level and instance-level.

\begin{wraptable}{r}{0.3\linewidth}
    \vspace{-25pt}
  \centering
  \scriptsize
  \setlength{\tabcolsep}{1.0mm}
  \caption{Comparison of different SSL approaches.}
    \vspace{-5pt}
    \resizebox{0.99\linewidth}{!}{
    \begin{tabular}{ lc }
\toprule
 Method & Accuracy \\
\midrule
 Point-MAE~\cite{pointmae} & 73.32  \\
 Linear probing &  71.09\\
 \midrule
  + FixMatch~\cite{fixmatch}&  73.48 \\
  + FlexMatch~\cite{flexmatch} & 75.51  \\
  + Dash \cite{dash} & 74.81 \\
  +  Chen \textit{et al.}\cite{chen2021consistency}  & 74.37 \\
    + M2CP \cite{chen2021multimodal}  & 79.88\\
    + Confid \cite{chen2023class}  & 81.24 \\
    + AllMatch \cite{paul2024improving} & 82.62\\
  \rowcolor{linecolor!40}+ \textbf{MCFT w. AllMatch}  & \textbf{84.49} \\
\bottomrule
\end{tabular}
}
 \vspace{-20pt}
  \label{tab:ssl2}
\end{wraptable}
\textbf{Semi-supervised approach.} Table \ref{tab:ssl_fewshot} compares the semi-supervised variant of our proposed method against state-of-the-art methods on the ModelNet40 dataset in the 5-shot setting \textit{(full few-shot)}. Here, only a small portion of labelled data (5 samples per class) is available, along with the remaining data as the unlabelled set. We report the average accuracy as well as the standard deviation over 3 runs. The semi-supervised variant of MCFT shows significant improvement (\textit{84.49\%}) over the fully-supervised variant and improved existing SOTA by 9.43\%, showing the positive impact of utilizing unlabelled data in tuning a foundation model under few-shot setup. 

In Table \ref{tab:ssl2}, we further compare the performance of MCFT(SSL) with different semi-supervised methods from various domains on the ModelNet40 dataset. For a fair comparison, we reproduce the performance of different SSL methods under the 5-shot setting with the PointMAE encoder and report the overall accuracy. Here, MCFT continues to gain a substantial improvement over all methods.

\begin{wraptable}{r}{0.35\linewidth}
\vspace{-20pt}
  \centering
  \scriptsize
  \setlength{\tabcolsep}{0.5mm}
  \caption{Performance of the pruned variant of MCFT under semi-supervised setting.}
    \vspace{-5pt}
    \resizebox{0.99\linewidth}{!}{
    \begin{tabular}{lcc}
    \toprule
       {Methods} & \# TP (M) &  5-shot \\
    \midrule
    Point-MAE~~\cite{pointmae} & 22.1 & 73.32$\pm$2.1 \\
   + IDPT~\cite{idpt}  & 23.8 & 73.51 $\pm$2.8\\
   + DAPT~\cite{dapt}  & 23.2 & 73.74$\pm$1.6 \\
   + PointGST~\cite{pointgst} &  22.7 & 75.06$\pm$1.9 \\
   \rowcolor{linecolor!40}+ MCFT & 22.7 & 78.36$\pm$\textbf{0.5} \\
   \rowcolor{linecolor!40}+ MCFT (SSL) & 22.7  & \textbf{84.49}$\pm$0.7 \\
   \rowcolor{linecolor!40}+ \textbf{MCFT (pruned)} & \textbf{20.9} & \textit{83.51}$\pm$1.1 \\
    \bottomrule
    
    \end{tabular} 
    }
  \label{tab:prune}
  \vspace{-20pt}
\end{wraptable}
\textbf{Pruning approach.} As discussed in the method section, structural pruning selectively removes less significant encoder layers to improve computational efficiency. Table \ref{tab:prune} shows the effect of pruning on the semi-supervised MCFT variant under the 5-shot setup. The compressed MCFT outperforms existing SOTA methods by 8.45\%. Additionally, it surpasses our supervised variant by 5.15\% with 1.8M fewer parameters in total. This light-weight version of MCFT maintains strong performance while reducing computational overhead, making it well-suited for resource-constrained applications.

\textbf{Inference Time Analysis.} 
Table \ref{tab:inference_time} compares different fine-tuning methods for 3D foundation models based on trainable parameters, computational cost (FLOPs), processing efficiency (throughput), and performance (PB\_T50\_RS). MCFT achieves the highest performance (87.2\%) while maintaining efficiency, with 22.7M parameters and 4.8G FLOPs, comparable to Point-MAE but outperforming it. MCFT (pruned) further reduces parameters (20.9M) and FLOPs (4.4G), significantly improving throughput (350.75 frames/s) but with a slight performance drop (83.5\%), making it ideal for resource-constrained scenarios.

\begin{wraptable}{r}{0.55\textwidth}
\footnotesize
\vspace{-25pt}
\setlength{\tabcolsep}{0.5mm}
\centering
\caption{Comparison of inference efficiency on object recognition task on ScanObjectNN. We measure throughput with a batch size of 32 on a single RTX 4090 GPU.}
\label{tab:inference_time}
\resizebox{0.99\linewidth}{!}{
\begin{tabular}{ lcccc }
\toprule
 Method & $\#$TP(M) & FLOPs(G) & Throughput (frame/s) & PB\_T50\_RS \\
 
\midrule
 Point-MAE~\cite{pointmae} & 22.1 & 4.8 & 323.66  & 85.2 \\
  IDPT~\cite{idpt}  &  23.8 & 7.2 & 281.18 & 84.9 \\
  DAPT \cite{dapt}  & 23.2 & 5.0 & 311.28 & 85.1 \\
  \rowcolor{linecolor!40}+ \textbf{MCFT}   & 22.7 & 4.8 & 321.58 & 87.2 \\
  \rowcolor{linecolor!40}+ \textbf{MCFT (pruned)}   & 20.9 & 4.4 & 350.75 & 83.5 \\

\bottomrule
\vspace{-35pt}
\end{tabular}
}
\end{wraptable}
\textbf{Sensitivity study. }
We conduct a comprehensive sensitivity analysis on MCFT using ModelNet40 to evaluate the impact of key hyperparameters. The EMA value plays a critical role in balancing adaptation and stability, with EMA $=0.999$ achieving the best performance, while lower values risk overfitting and higher values limit learning. Pruning analysis shows that removing one encoder layer incurs only a minor accuracy drop, whereas aggressive or excessive pruning leads to sharp degradation, highlighting the importance of gradual pruning. The pruning schedule is also influential, where pruning one layer every $K=10$ epochs yields the best accuracy and stability compared to more frequent or delayed schedules. Additionally, the model is relatively robust to variations in the loss factor, achieving optimal performance at a value of 1.0. Finally, delaying full fine-tuning beyond 50 epochs negatively impacts accuracy, indicating that an early transition to full fine-tuning is crucial for optimal performance.

\vspace{-5pt}
\section{Conclusion}
\label{conclusion}
\vspace{-5pt}

We introduced Momentum-Consistency Fine-Tuning (MCFT), an adapter-free approach for adapting 3D foundation models in data-scarce settings. By selectively fine-tuning the encoder while enforcing momentum-based consistency, MCFT effectively mitigates overfitting and representation drift commonly observed in full fine-tuning, without incurring the inference-time overhead of PEFT methods. This positions MCFT as a practical middle ground between full and parameter-efficient fine-tuning. 
Extensive evaluations on object recognition and part segmentation tasks show that MCFT consistently outperforms existing approaches across few-shot, full few-shot, and fully supervised setups, achieving a notable 3.30\% improvement in the 5-shot regime. We further demonstrate that leveraging unlabeled data through a semi-supervised extension yields additional performance gains, while a structured pruning variant maintains comparable accuracy with reduced computational cost, making MCFT suitable for deployment on resource-constrained devices. 
While MCFT is effective, its performance depends on the choice of momentum and pruning hyperparameters, which may require dataset-specific tuning. Future work will explore adaptive strategies for hyperparameter selection and extend MCFT to more complex 3D understanding tasks, including large-scale scene-level reasoning.

%
%
\bibliographystyle{splncs04}
\bibliography{ref}
\end{document}